# Etat de l'art des systèmes robotisés en vue d'une application pour la chirurgie otologique

**M. Dreux[a], L. Ginzburg[a], P. Bordure[b], D. Chablat[c], G. Michel[b,c]**

a. Ecole Centrale de Nantes,
1 rue de la Noë, 44321 Nantes
{manon.dreux@eleves.ec-nantes.fr, lea.ginzburg@eleves.ec-nantes.fr}
b. CHU de Nantes, 5 Allée de l'Île Gloriette, 44093 Nantes
{philippe.bordure@chu-nantes.fr, guillaume.michel@chu-nantes.fr}
c. Laboratoire des Sciences du Numérique de Nantes (LS2N), UMR CNRS 6004,
1 rue de la Noë, 44321 Nantes
damien.chablat@cnrs.fr

## Résumé :

*L'objectif de cet article est la réalisation d'un état de l'art des brevets en lien avec la robotisation des porte-endoscopes utilisés en otologie. Dans une première partie, nous présentons les fonctions d'un endoscope ainsi que l'environnement dans lequel nous souhaitons l'utiliser : l'oreille moyenne. Dans un second temps, nous avons réalisé un état de l'art des brevets existant dans les domaines de la robotique et de la chirurgie otologique. Cette approche se base sur une méthodologie précise de recherche qui utilise soit des mots-clés, soit la classification des brevets. A partir de ce résultat, nous avons classifié les brevets en fonction de leurs pays de dépôt et de leur date de publication afin d'en faire une synthèse. Cette étude permet de connaître les solutions techniques introduites ces dernières années afin de guider notre démarche d'innovation.*

## Abstract:

*This article deals with a patents state-of-the-art linked to our robotic system which holds an endoscope. In a first part, we analyze how the endoscope operates as well as we discover the environment where we wish to use it: the middle ear. In a second step, we establish a state-of-the-art on patents existing in the robotic and surgical fields. The approach is based on a specific search methodology using keywords and classifications. Thanks to all the patents found, we gather them according their country and date of publication, and we are able to analyze the results. This study brings to light technical solutions invented in recent years and allows us to find an innovative research axis.*

**Mots-clés : brevets, robotique, chirurgie otologique, endoscope**

# 1    Introduction

Lors d'une opération ORL (Oto-Rhino-Laryngologie), les mains du chirurgien sont prises par les outils médicaux. Ainsi, les chirurgiens sollicitent de plus en plus la création d'un robot aidant à la visualisation interne de l'oreille remplaçant une vision externe par lunettes binoculaires. Jusqu'à récemment, il n'existait pas de dispositif répondant à ce besoin en toute sécurité. Une solution à ce problème consiste en un système robotisé de porte endoscope, automatisé à l'aide d'un contrôle par suivi des gestes du chirurgien ou simplement à l'aide d'un joystick. L'objectif de notre projet est de réaliser un état de l'art le plus complet possible dans le domaine de la robotique médicale. Il s'agit donc de comparer l'architecture envisagée avec les brevets trouvés afin d'éviter de reproduire des mécanismes existants.

# 2    Endoscope médical

Le point de départ de cet étude consiste en la recherche de brevets pour élaborer un nouveau système robotisé porte-endoscope. Une première étape a donc été tout naturellement de faire quelques recherches préalables sur les endoscopes.

L'endoscope est un instrument optique médical permettant la réalisation d'une endoscopie, c'est-à-dire la visualisation d'opérations chirurgicales intrusives. Il est composé d'un tube muni d'un système d'éclairage et d'une petite caméra vidéo à l'extrémité du tube qui va filmer et retransmettre les images de haute qualité sur un écran.

Afin de réaliser des recherches sur les endoscopes, il est important de comprendre l'environnement dans lequel nous voulons l'utiliser. Dans le cadre de notre projet, le chirurgien intervient au niveau de l'oreille externe et de l'oreille moyenne (**Figure 1**).

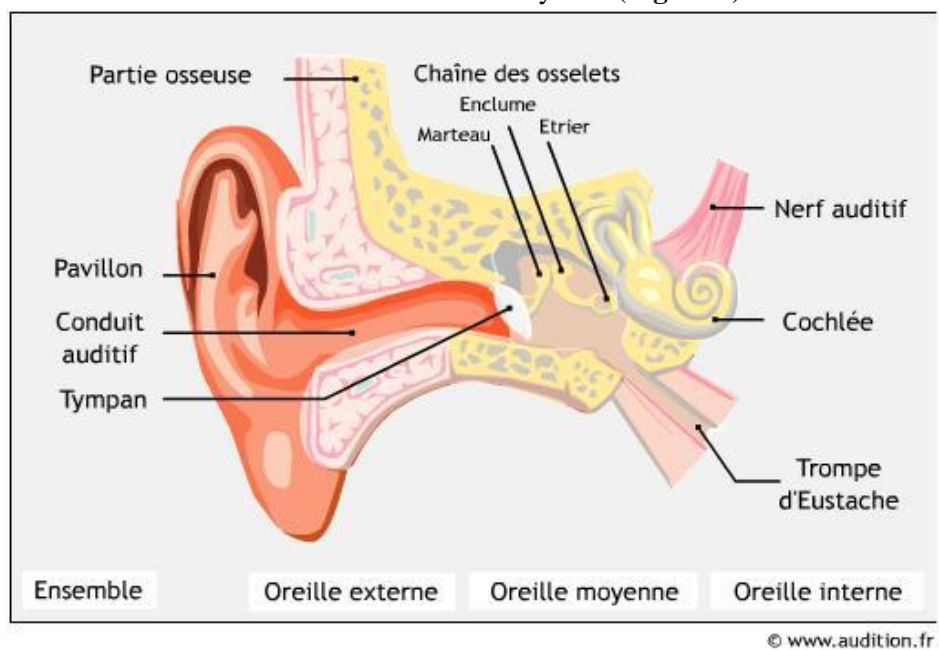

**Figure 1 : Anatomie de l'oreille**

Il faut noter la difficulté de l'opération puisqu'elle implique de travailler dans une zone très étroite avec des dimensions très petites. En effet, le marteau est de l'ordre de 8mm de long, l'enclume de 6mm et l'étrier de 3mm. Le nerf facial traverse également l'oreille moyenne. Le chirurgien, ainsi que le bras robotisé étudié, doivent préserver ce nerf lors d'une opération pour éviter une paralysie faciale du patient.

L'objectif du projet global est de remplacer la vision par microscope en une vision par endoscope. En effet, le microscope renvoie une image assez éloignée de l'oreille tandis que l'endoscope permet d'obtenir une meilleure vue en insérant la caméra dans l'oreille moyenne, juste au niveau de la zone d'opération. L'endoscope permet ainsi au chirurgien de voir des zones inaccessibles avec le microscope. Ce qui facilite l'opération dans l'espace restreint de l'oreille moyenne. Malheureusement, actuellement le chirurgien utilise déjà ses deux mains pour manipuler un outil chirurgical et un outil d'aspiration, d'où l'intérêt d'un robot porte-endoscope. Actuellement, les mouvements réalisés par le chirurgien sont principalement des rotations autour d'un point fixe situé dans le conduit auditif ainsi que des translations le long de ce conduit.

## 3    Méthodologie de recherches de brevets

Dans l'optique de la création d'un robot porte-endoscope, il est nécessaire de réaliser une veille sur les brevets, c'est-à-dire de rechercher les brevets déjà déposés sur des systèmes proches de celui du projet, ou possédant des caractéristiques similaires. Cette démarche est nécessaire si l'on souhaite déposer un brevet à la fin de notre étude.

Nous devons notamment nous intéresser au caractère public ou privé des brevets et aux pays dans lesquels ils s'appliquent si notre étude utilise une partie d'un brevet existant. En effet, un brevet peut rester dans le domaine privé jusqu'à 20 ans après sa publication si les taxes annuelles sont bien payées. Lorsqu'un inventeur utilise tout ou partie d'un brevet toujours privé, il devra payer des taxes auprès de l'auteur de celui-ci. En revanche, lorsqu'un brevet tombe dans le domaine public, l'invention peut être utilisée sans démarche particulière.

Il existe différentes banques de données de brevets. Nous avons opté pour Espacenet [1] même si d'autres sites web offrent des services similaires comme Google Patents [7]. Il s'agit de services gratuits regroupant des brevets du monde entier et proposant une recherche avancée où l'on peut préciser des mots clés, des numéros de publication ou encore des noms d'auteurs.

Nous avons principalement effectué deux types de recherches :
- par mots clés
- par classification.

Les mots clés utilisés qui sont issus du monde médical et de la robotique sont : RCM (Remote Center of Motion), poignet sphérique (Spherical Wrist), chirurgie robotique (Robotic Surgery), parallèle (Parallel) pour l'architecture parallèle du robot. Nous avons principalement utilisé les classifications présentées dans le **Tableau 1**.

| A61B34/70 | Manipulateurs spécialement adaptés pour l'utilisation en chirurgie |
|---|---|
| A61B34/20 | Systèmes de navigation chirurgicale ; dispositifs de suivi ou de guidage d'instruments chirurgicaux |
| A61B34/30 | Robots chirurgicaux |
| A61B1/00 | Instruments pour l'examen médical de l'intérieur de cavités ou de tubes du corps par inspection visuelle ou photographique, exemple : endoscopes |

Tableau 1 : Classifications des brevets

Ainsi, nous avons combiné des recherches en mécanique pure et en chirurgie pour étendre au maximum nos recherches. Nous avons mis en évidence 24 brevets. Evidemment, il ne s'agit que d'un aperçu de ce que l'on peut trouver en rapport avec notre projet, puisqu'il existe par exemple des recherches non brevetées.

## 4    Etat de l'art des brevets

Nous regroupons dans le tableau ci-dessous les 24 brevets trouvés lors de nos recherches. Les références de ceux-ci permettent de les retrouver rapidement afin d'avoir plus de détails sur leur contenu.

Le **tableau 2** regroupe les architectures de robots parallèles permettant de réaliser un mouvement sphérique. Ces mécanismes peuvent être utilisés pour orienter un endoscope dans le conduit auriculaire. On remarque que ces brevets sont anciens et donc dans le domaine public. Il faut noter que le centre de rotation est soit un point virtuel, c'est-à-dire l'intersection d'axes de rotation, soit un cardan ou une rotule pour fixer le centre de rotation [6].

| **Référence brevet** | **Nom** | **Date** | **Image** |
|---|---|---|---|
| CA2235759 | TWO DEGREE-OF-FREEDOM SPHERICAL ORIENTING DEVICE | 1998-10-23 | 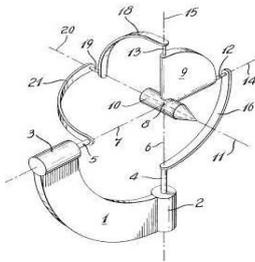 |
| US4628765 | SPHERICAL ROBOTIC WRIST JOINT | 1986-12-16 | 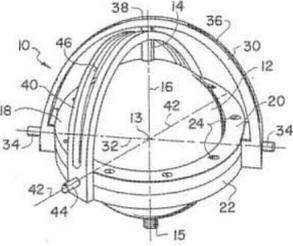 |
| US5243873 | TWO-AXIS MOTION MECHANISM | 1993-09-14 | 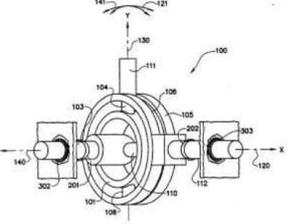 |
| US4878393 | DEXTROUS SPHERICAL ROBOT WRIST | 1989-11-07 | 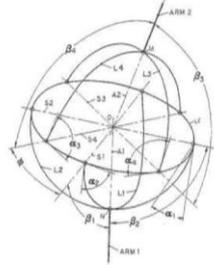 |

Tableau 2 : Robots sphériques

Pour les mécanismes RCM présentés dans le **Tableau 3**, c'est-à-dire centre de mouvement distant, le centre de rotation du mouvement peut être défini de trois manières différentes:
- Soit par logiciel
- Soit par des contraintes géométriques issues des intersections des axes de rotation des actionneurs
- Soit par des parallélogrammes ou une synchronisation des mouvements avec des courroies.

La première solution permet la génération de plusieurs types de mouvement mais offre un degré de sécurité moindre. La deuxième solution génère un grand volume de débattement du bras pour "tourner" autour du centre de rotation. La contrepartie peut être un espace de travail réduit si l'on souhaite diminuer ce défaut. Finalement, la troisième solution permet de déplacer tout ou partie de l'actionnement loin du centre de rotation. Il est à noter que tous ces mécanismes peuvent être utilisés pour des applications médicales pour porter des outils et plus rarement pour faire de l'imagerie.

| **Référence brevet** | **Nom** | **Date** | **Image** |
|---|---|---|---|
| WO2017055990 | OPTICAL REGISTRATION OF A REMOTE CENTER OF MOTION ROBOT | 2017-04-06 | |
| WO2014020571 | CONTROLLER DEFINITION OF A ROBOTIC REMOTE CENTER OF MOTION | 2014-02-06 | |
| US2007173977 | CENTER ROBOTIC ARM WITH FIVE-BAR SPHERICAL LINKAGE FOR ENDOSCOPIC CAMERA | 2007-07-26 | |
| WO2008157225 | ROBOTIC MANIPULATOR WITH REMOTE CENTER OF MOTION AND COMPACT DRIVE | 2008-12-24 | |
| FR2845889 | SURGICAL ROBOT FOR GUIDING AND POSITIONING AN INSTRUMENT HAS SUPPORTING LEG WITH ROTARY BEAM WITH DRIVE, CARRIER ARM AND DEFORMABLE PARALLELOGRAMS | 2004-04-23 | |
| WO2017192796 | REMOTE CENTER OF MOTION ROBOT | 2017-11-09 | 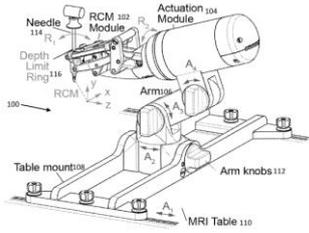 |
| WO2012065058 | REMOTE CENTER OF MOTION ROBOT FOR MEDICAL IMAGE SCANNING AND IMAGE-GUIDED TARGETING | 2012-05-18 | 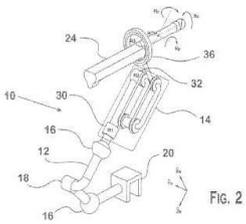 |
| CA2475239 | REMOTE CENTER OF MOTION ROBOTIC SYSTEM AND METHOD | 2003-08-14 | 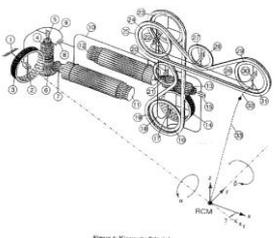 |
| KR20090089928 | CURVED RCM OF SURGICAL ROBOT ARM | 2009-08-25 | 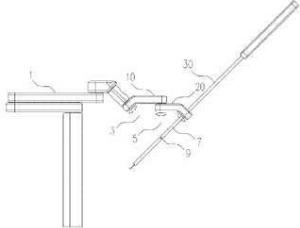 |

**Tableau 3 : Robots avec RCM**

Pour les robots chirurgicaux présentés dans le **Tableau 4**, il faut distinguer les architectures sur base mobile et les architectures fixes. Ces brevets portent soit sur le support d'outil soit sur leur intégration complète. C'est la société Intuitive Surgical Operations qui est à l'origine du brevet initial utilisé dans le robot Da Vinci déposé en octobre 1999 (**Figure 2**).

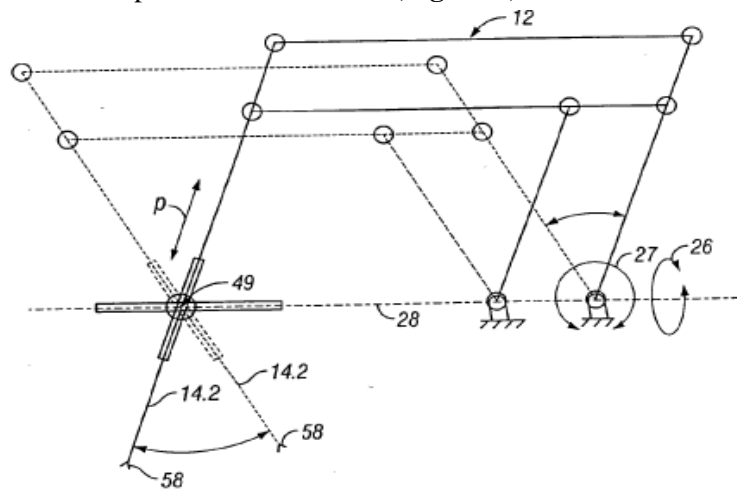

**Figure 2 : RCM issu du brevet US20070038214**

| Référence brevet | Nom | Date | Image |
|---|---|---|---|
| US2014128882 | SURGICAL INSTRUMENT, SUPPORT EQUIPMENT, AND SURGICAL ROBOT SYSTEM | 2014-05-08 | 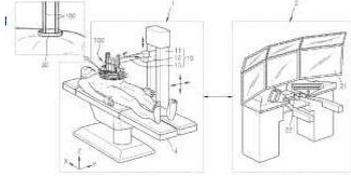 |
| US20070038214 | MINIMALLY INVASIVE SURGICAL HOOK APPARATUS | 1999-10-08 | 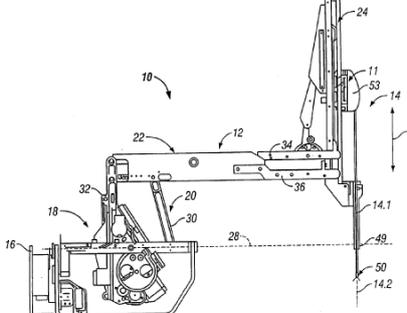 |
| US2017189128 | METHODS, SYSTEMS, AND DEVICES FOR MOVING A SURGICAL INSTRUMENT COUPLED TO A ROBOTIC SURGICAL SYSTEM | 2017-07-06 | 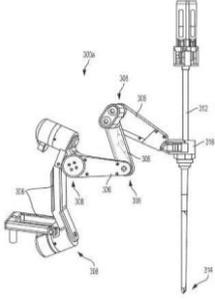 |
| US2018318021 | SYSTEM AND METHOD FOR IMAGE-BASED ROBOTIC SURGERY | 2014-07-03 | 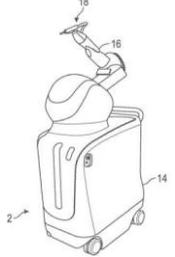 |
| WO2017103984 | MEDICAL MANIPULATOR SYSTEM AND OPERATING METHOD THEREOF | 2017-06-22 | 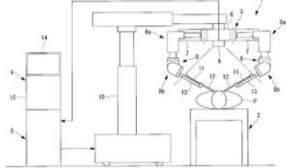 |

Tableau 4 : Robots chirurgicaux

Le **Tableau 5** présente une liste d'instruments chirurgicaux qui ont été montés sur des architectures robotisées avec une structure sérielle. On peut noter que ces dispositifs portent soit un outil, soit un endoscope. La différence essentielle vient du dimensionnement et de la précision, car une caméra n'applique pas d'effort sur son environnement (normalement, elle ne touche pas le patient) et les mouvements de la caméra sont de grande amplitude mais à faible vitesse. Le chirurgien cherche surtout une image stable pour éviter une naupathie due aux mouvements de la caméra.

| Référence brevet | Nom | Date | Image |
|---|---|---|---|
| US2018317753 | ENDOSCOPIC SYSTEM AND METHOD FOR CONTROLLING THE SAME | 2018-11-08 | 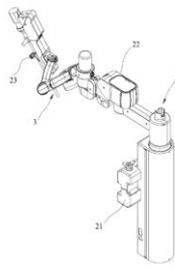 |
| US2013317517 | DISPOSITIF D'ASSISTANCE A LA CHIRURGIE OTOLOGIQUE D'UN PATIENT A IMPLANTER AVEC UN IMPLANT COCHLEAIRE | 2013-11-28 | 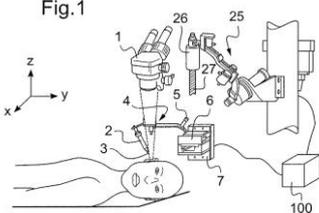 |
| FR3066378 | INSTRUMENT CHIRURGICAL A PORTIONS DEVIEES ET INSTALLATION ROBOTISEE COMPORTANT UN TEL INSTRUMENT | 2018-11-23 | 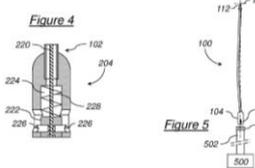 |

**Tableau 5 : Instruments chirurgicaux avec structure sérielle**

D'autres architectures de robots parallèles, notées dans le **tableau 6**, peuvent être utilisées pour porter des outils ou des endoscopes mais qui n'utilisent pas la notion de centre de rotation déporté.

| Référence brevet | Nom | Date | Image |
|---|---|---|---|
| DE102017111296 | ROBOTIC MANIPULATOR FOR GUIDING AN ENDOSCOPE, HAVING PARALLEL KINEMATICS | 2018-08-02 | 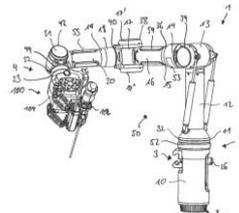 |
| CN107363809 | FOUR-DEGREE-OF-FREEDOM PARALLEL TYPE MINIMALLY INVASIVE SURGERY | 2017-11-21 | 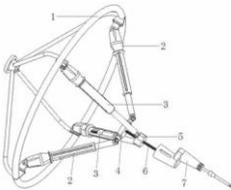 |
| WO2012013846 | PARALLEL-CONFIGURATION TELEROBOTIC ARM FOR USE IN MINIMALLY INVASIVE SURGERY | 2012-02-02 | 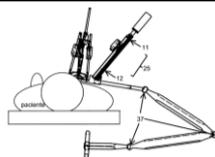 |

**Tableau 6 : Instruments chirurgicaux avec structure parallèle**

# 5 Discussion et Conclusion

Cette étude nous a permis d'identifier les architectures de robots pouvant être utilisées pour créer un robot sphérique, un centre de rotation déporté et pour porter un outil ou un endoscope. Certains de ces robots ont déjà comme vocation une utilisation en chirurgie mais peu connaissent une utilisation commerciale importante. Nous avons classé l'ensemble des brevets selon leurs pays de publication dans la **Figure 3**.

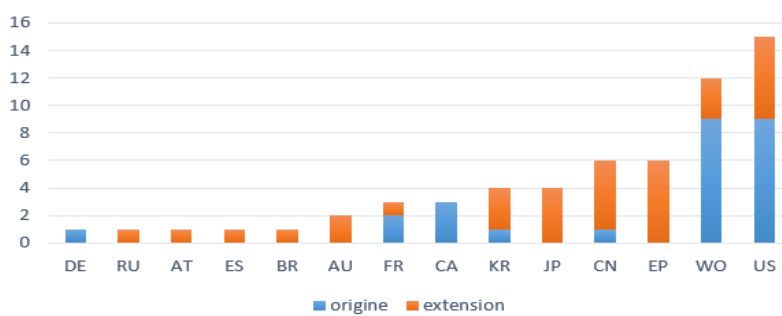

Figure 3 : Nombre de brevets en fonction des pays de publication

On peut classer les pays en fonction de l'importance du nombre de brevets déposé : Etats-Unis, monde, Canada, France, Allemagne, Corée et Chine. Puis les extensions sont réalisées dans les autres pays. Concernant les extensions, elles ont été principalement réalisées en Europe, en Chine et aux Etats-Unis.

Dans la **Figure 4**, nous avons ensuite classé l'ensemble des brevets selon leur année de dépôt. Nous pouvons constater que les brevets concernant le poignet sphérique ont majoritairement été déposés avant l'an 2000. Ceci révèle que cette invention est relativement ancienne. En revanche, en ce qui concerne les mécanismes parallèles, les instruments de chirurgie et la robotique dans la médecine, les brevets sont récents. De même, pour le RCM, les brevets ont été déposés après l'an 2000. Ainsi, cet histogramme révèle que le système étudié de porte-endoscope est un sujet d'actualité et qu'il répond à un véritable besoin de la part des hôpitaux.

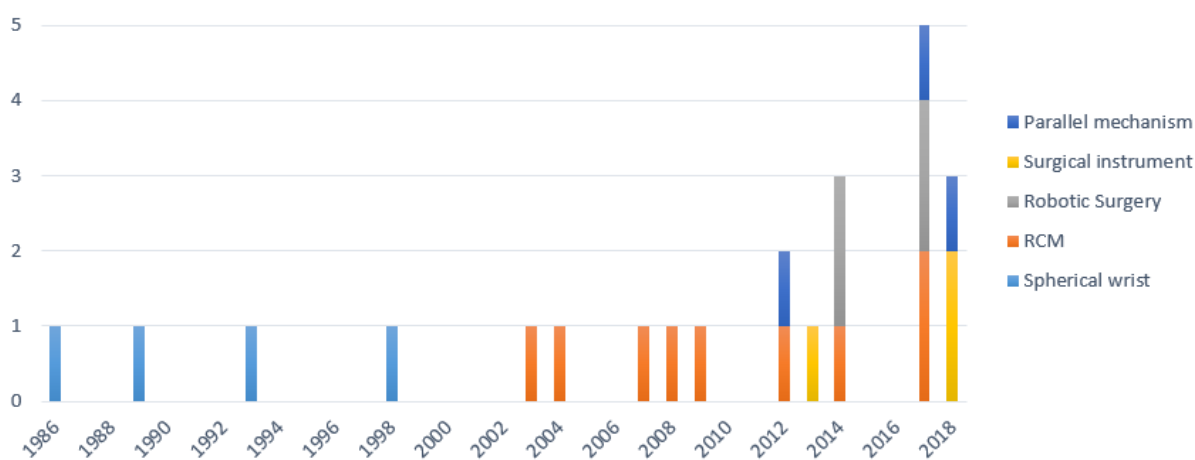

Figure 4 : Nombre de brevets en fonction des années de dépôt

Ainsi, les différentes recherches sur les endoscopes et les inventions en mécanique, robotique et chirurgie ont permis de préciser l'architecture du robot. Ce travail constitue un premier pas vers la publication d'un nouveau brevet sur un système robotisé pour la chirurgie otologique qui fera l'objet d'une prochaine publication. Par ailleurs, l'utilisation d'un tel système est envisageable pour d'autres applications comme les opérations des sinus.

## Références

[1] Office Européen des brevets. Espacenet. *Patent Search*. [en ligne], https://worldwide.espacenet.com/advancedSearch?locale=en_EP, consulté en Avril 2019.

[2] G. Kazmitcheff. Modélisation dynamique de l'oreille moyenne et des interactions outils organes pour la conception d'un simulateur appliqué à la chirurgie otologique, Informatique [cs], Thèse, Université des sciences et des technologies de Lille 1, 2014.

[3] M. Miroir. Conception d'un système robotisé d'aide à la microchirurgie otologique: application au traitement de l'otospongiose, Thèse, Université Paris 6, 2009.

[4] M. Miroir, Y. Nguyen, J. Szewczyk, O. Sterkers, A. Bozorg Grayeli, Design, kinematic optimization, and evaluation of a teleoperated system for middle ear microsurgery, Scientific World Journal, 2012.

[5] M. Badr-El-Dine, A.L. James, G. Panetti, D. Marchioni, L. Presutti, J.F. Nogueira. Instrumentation and technologies in endoscopic ear surgery, Otolaryngologic Clinics of North America, 2013, Volume 46, Issue 2, 211 - 225

[6] S.K. Agrawal, G. Desmier, and S. Li, Fabrication and analysis of a novel 3 dof parallel wrist mechanism, *ASME J. of Mechanical Design*, 117(2):343-345, Juin 1995

[7] Google Patent [en ligne], https://www.google.com/?tbm=pts, consulté Avril 2019.